\documentclass[11pt,a4paper]{article}
\usepackage[hyperref]{acl2021}
\usepackage{times}
\usepackage{latexsym}

\usepackage{microtype}

\usepackage{graphicx}
\usepackage{multirow}
\usepackage{multicol}
\usepackage{tablefootnote}
\usepackage{booktabs}

\aclfinalcopy 

\newcommand{\printfnsymbol}[1]{%
  \textsuperscript{\@fnsymbol{#1}}%
}

\newcommand*\samethanks[1][\value{footnote}]{\footnotemark[#1]}

\title{SemEval-2021 Task 9: Fact Verification and Evidence Finding for Tabular Data in Scientific Documents (SEM-TAB-FACTS)}

\makeatletter
\def\@fnsymbol#1{\ensuremath{\ifcase#1\or *\or
   \mathsection\or \mathparagraph\or \|\or **\or \dagger\dagger
   \or \ddagger\ddagger \else\@ctrerr\fi}}

\author{Nancy X. R. Wang\thanks{\quad Equal Contribution} \quad Diwakar Mahajan\samethanks \quad Marina Danilevsky \quad Sara Rosenthal\thanks{\quad Corresponding Author}\\
IBM Research\quad \\
nancywang1991@gmail.com, \{dmahaja, mdanile, sjrosenthal\}@us.ibm.com}

\date{}

\begin{document}
\maketitle
\begin{abstract}
Understanding tables is an important and relevant task that involves understanding table structure as well as being able to compare and contrast information within cells. In this paper, we address this challenge by presenting a new dataset and tasks that addresses this goal in a shared task in SemEval 2020 Task 9: Fact Verification and Evidence Finding for Tabular Data in Scientific Documents (SEM-TAB-FACTS). Our dataset contains 981 manually-generated tables and an auto-generated dataset of 1980 tables providing over 180K statement and over 16M evidence annotations. SEM-TAB-FACTS featured two sub-tasks. In sub-task A, the goal was to determine if a statement is supported, refuted or unknown in relation to a table. In sub-task B, the focus was on identifying the specific cells of a table that provide evidence for the statement. 69 teams signed up to participate in the task with 19 successful submissions to subtask A and 12 successful submissions to subtask B. We present our results and main findings from the competition. 
\end{abstract}

\section{Introduction}

\begin{figure}
    \centering
    \includegraphics[width=\columnwidth ]{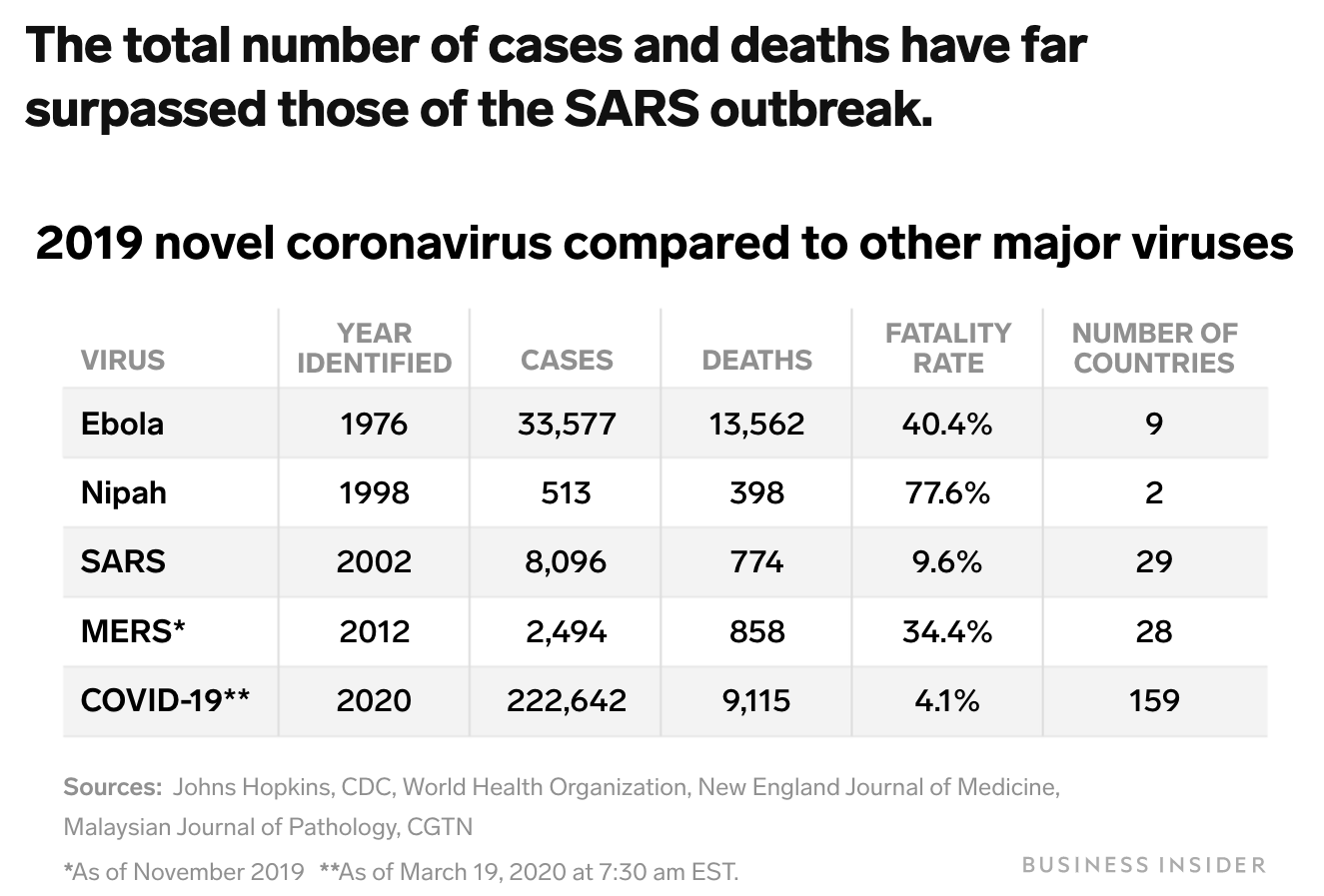}
    \caption{Surrounding text often highlights some information from the table but does not capture all data. Alternately, the linked text may be subjective or even misleading without the original table to check the claims.}
    \label{fig:covid_table}
\end{figure}

Tables are ubiquitous in documents and presentations for conveying important information in a concise manner. This is true in many domains, stretching from scientific to government documents. In fact, surrounding text in these articles are often statements summarizing or highlighting some information derived from the primary source of data in tables. A relevant example is shown in Figure~\ref{fig:covid_table} from a Business Insider article analyzing the impact of Covid-19~\cite{covidArticle}. Describing all the information provided in this table in a readable manner would be lengthy and considerably more difficult to understand. 
Despite their importance, popular question answering (e.g. SQuAD and Natural Question \cite{rajpurkar2016squad, 47761}) and truth verification tasks (e.g. SemEval-2019 Fact Checking Task \cite{mihaylova-etal-2019-semeval}) have not focused on tables, being composed solely of written text.
This is likely due to their complexity to parse and understand, despite their rich amount of information. 

\begin{figure}
    \centering
    \includegraphics[width=\columnwidth, trim=20 122 300 30, clip]{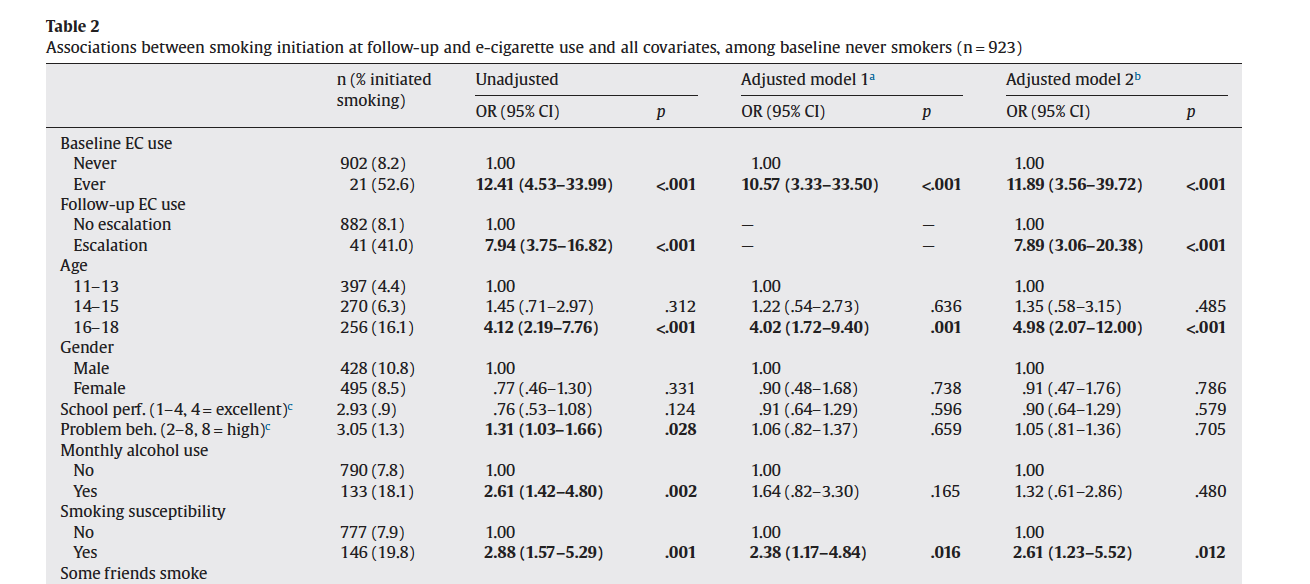}
    \caption{A complex table sourced from~\cite{east2018association} with hierarchical column and row structure. Additional difficulty follows from row hierarchy not being delineated by separate columns.}
    \label{fig:complex_table}
\end{figure}

Further, the structure of tables allows much more information to be presented in an efficient manner as humans can interpret meaning in the spatial relationship between cells. However, due to their challenging nature, recent algorithms have been less successful at extracting~\cite{2019-table-repair} and understanding header and data structure in tables~\cite{cafarella2018ten}. In addition, any hierarchical and nested headers (common in printed documents) increases the difficulty in interpreting data cells, as shown in Figure~\ref{fig:complex_table}.

In this paper, we propose to bridge this gap with statement verification and evidence finding using tables from scientific articles. This important task promotes proper interpretation of the surrounding article. In fact, the misunderstanding of tables can lead to the reporting of fake news that we see as being all too prevalent today.

We present the first SemEval challenge to address table understanding. We introduce a brand new dataset of 1980 tables from scientific articles that addresses two challenging tasks important to table understanding:

\begin{description}
    \setlength\itemsep{-0.1em}
    \item[A: Statement Fact Verification] Given a statement, determine whether it is \textit{supported}, \textit{refuted} or \textit{unknown} according to the table. 
    \item[B: Cell Evidence Selection] Given a statement, select the cells in the table that provide evidence supporting or refuting the statement.
\end{description}

The rest of this paper is formatted as follows: We first discuss related work. We then present a new large table understanding dataset containing close to 2000 tables that is the first to provide evidence labels at the cell level for statements and the first to focus on scientific articles. We provide a detailed analysis of the dataset including several baseline results. We then discuss the performance and approaches of the 19 participants in our challenge and end with an aggregated analysis of participating teams. Finally, we discuss future work.



\section{Related Work}
\paragraph{Natural Language Inference (NLI)} The table evidence task can be best understood as a variation of the natural language inference task \cite{dagan2005}, but on tabular data. NLI asks whether one (or more) sentence entails, refutes, or is unrelated to another sentence; our framing asks whether a given table entails, refutes, or is unrelated to a sentence. Several datasets have been created for studying NLI, such as SNLI  \cite{bowman-etal-2015-large}, MultiNLI \cite{williams-etal-2018-broad}, and SciTail \cite{khot2018scitail}. 
\paragraph{Table QA}

This task is also closely related to the problem of search and question answering on tables. The closest example would be, given a table that is known to contain the relevant information, return cell values that answer a natural language question~\cite{pasupat-liang-2015-compositional}. A variation requires analyzing a collection of tables rather than a single one, along with the natural language question~\cite{Sun16_TableQA}. Two of the most recent works are TAPAS~\cite{herzig-etal-2020-tapas} and TaBERT~\cite{yin-etal-2020-tabert}, which jointly pre-train over textual and tabular data to facilitate table QA. However, such approaches have previously focused on traditional natural language questions (``What is the population of France?'') rather than inference statements (``France has the highest population in Europe''), which may be entailed, refuted or unknowable from the given table.

\paragraph{Related Datasets} The works closest to our dataset are TabFact~\cite{chen2020tabfact} and INFOTABS~\cite{gupta2020infotabs}. Both datasets were sourced from Wikipedia tables and contain hypothesis and premise pairs. TabFact has entailment and refute hypothesis types while INFOTABS has an additional ``neutral'' hypothesis category, much like our ``unknown'' statements. Both works show that neural models still lag far behind human performance for the fact checking task with tables. 

While both datasets have been great at kindling interest in fact verification with tabular data, our dataset differs in two key aspects. First, we source from scientific articles in a variety of domains rather than Wikipedia infoboxes. Scientific tables have very specialized vocabulary and can be more difficult to interpret. Additionally, scientific tables have much more complex structure, like hierarchical column and row headers, rendering the assumption that the first column/row is the header unhelpful. Finally, tables are often directly referenced in scientific text unlike Wikipedia tables that are generally stand-alone. This creates an opportunity to leverage natural statements that depict the original author's style and intent. The second key differentiator of SEM-TAB-FACTS is the accompanying evidence annotations. We believe the future of fact verification and AI in general will be in cooperation with humans rather than in replacement. Thus, it is essential that models are able to present explanations for decisions on the relationship between the statement and table by showing the most relevant cells in a potentially very large table. 

\begin{table*}
\small
\centering
\begin{tabular}{lrrrrrr}
\toprule
\textbf{Source} & \textbf{\#Tables} & \textbf{\#Entailed} & \textbf{\#Refuted} & \textbf{\#Unknown} & \textbf{\#Relevant} & \textbf{\#Irrelevant} \\
\midrule
Train Crowdsourced & 981 & 2,818 & 1,688 & 0 & 0 & 0 \\
\midrule
Train Auto-generated & 1,980 & 92,136 & 87,209 & 0 & 1,039,058 & 15,467,957 \\
\midrule
Development & 52 & 250 & 213 & 93 & 3,048 & 2,8495 \\
\midrule
Test & 52 & 274 & 248 & 131 & 3,458 & 26,724 \\
\bottomrule
\end{tabular}
\caption{\label{table:stats} Statistics for our SEM-TAB-FACTS dataset.}
\end{table*}

\section{Dataset Details}

Our dataset consists of two forms of generation: (1) a crowdsourced dataset, and (2) an auto-generated dataset. Table \ref{table:stats} presents the statistics of the dataset. We detail our dataset creation process in the following sections.

\subsection{Data extraction and preprocessing}
We sourced our tables from scientific articles belonging to active journals that are currently being published by Elsevier and are available on ScienceDirect\footnote{\url{https://www.elsevier.com/__data/promis_misc/sd-content/journals/jnlactive.xlsx}}. We utilized Elsevier ScienceDirect APIs\footnote{\url{https://dev.elsevier.com/sd_apis.html}} to scrape scientific articles which belong to this list, and satisfy the following criteria: (1) the article is open-access\footnote{\url{https://www.elsevier.com/open-access}}, (2) the article is available under ``Creative Commons Attribution 4.0 (CC-BY)'' user license\footnote{\url{https://www.elsevier.com/about/policies/open-access-licenses/user-licences}}, and (3) the article has at least one table. We downloaded 1,920 articles belonging to 722 journals which contained 6,773 tables. We further filtered out complicated tables (e.g. multiple tables in a single table) using hand-written rules to get a set of 2,762 candidate tables from 1,085 articles for annotation. We also extracted sentences mentioning the table within the scientific article as candidate statements, which are corrected and then labeled manually by the annotators.

\begin{figure*}[ht]
    \centering
    \includegraphics[width=\textwidth ]{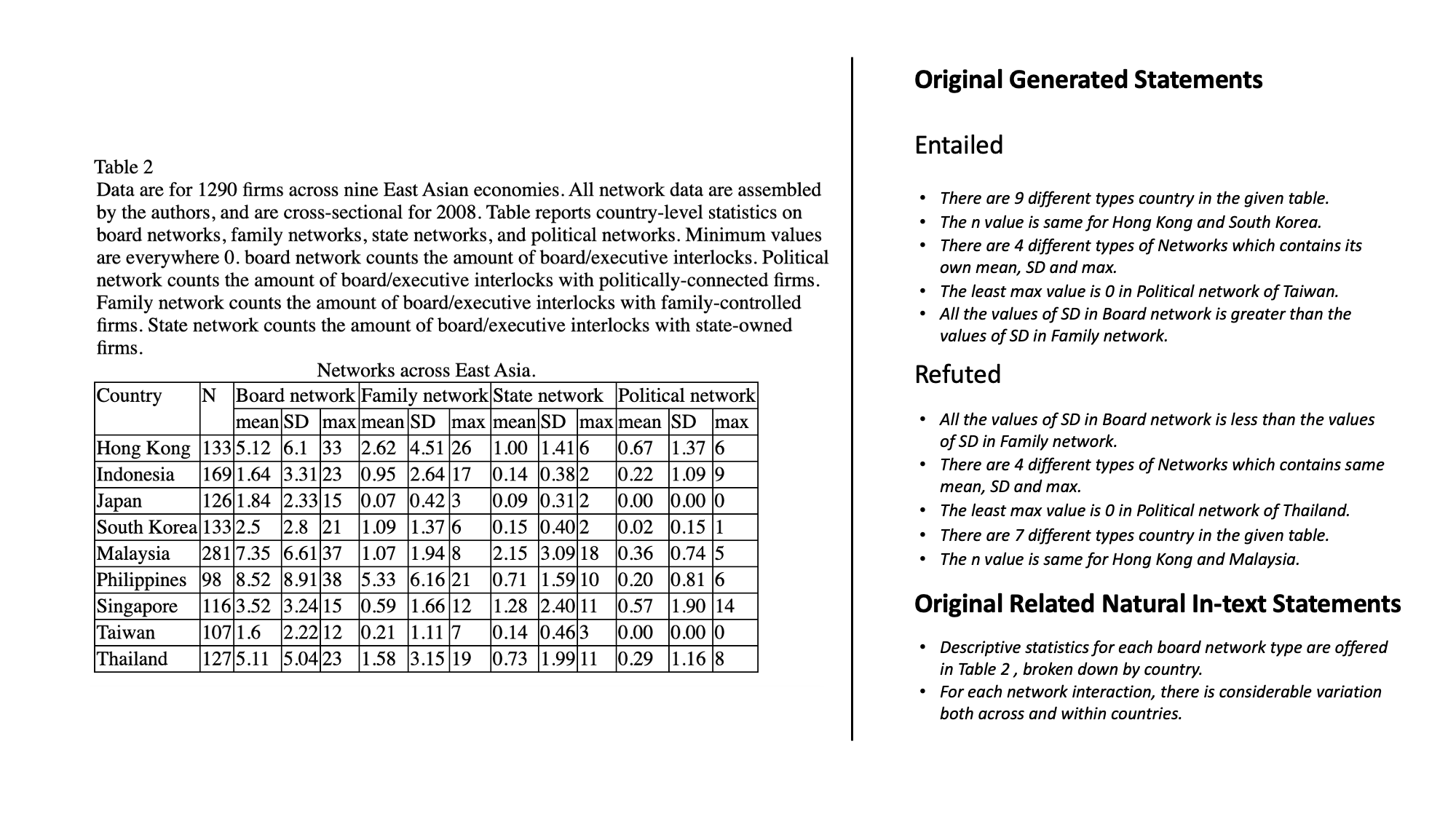}
    \caption{Sample crowd-sourced statements for one table (sourced from \cite{carney2020board}). Please note that these are the original statements without any further corrections nor rephrasing.}
    \label{fig:appen_table}
\end{figure*}

\subsection{Crowdsourced labeling}

The manually generated statements were collected using the crowdsourcing platform Appen\footnote{\url{https://appen.com/}}. We collected five entailed and five refuted statements for each table from the business preferred operators (BPO) on Appen. The BPO crowd is composed of employees hired by Appen on an hourly basis at a constant pay rate determined by Appen. We found that the workers were much more motivated for the task as they were able to ask questions if needed and we were also able to provide direct feedback to the workers. We initially attempted generating statements with workers from the Appen open-crowd, which is on-demand, but the quality was very poor as it was hard to automatically validate naturally generated statements.  Our instructions explicitly lay out 7 types of statements and ask that workers attempt to make one of each type. We encourage the use of different sets of cells whenever possible. The types of statements are aggregation, superlative, count, comparative, unique, all and usage of caption or common sense knowledge. These are derived from the INFOTABS analysis~\cite{gupta2020infotabs}. We asked workers to avoid subjective adjectives like ``best'', ``worst'', ``seldom'' and look-up statements that only require reasoning with one cell. The pay for each statement set was 75 cents. In total, we collected 10000 statements for 1000 unique tables. See Figure~\ref{fig:appen_table} for an example table with its manually generated and natural statements. 

Additionally, for our training data, we conducted a verification task to check for grammatical issues and doubly verify the statement label for both the generated and natural in-text statements. The verification task was paid at 3 cents per statement, which equates to 30 cents per table. We restricted the verification task to the workers in the open-crowd from English speaking countries. After verification, we only preserved the statements that were verified to be grammatically correct and the new label matched the original label. Natural statements were also verified in the same process. Although natural statements were generally factually correct, they were sometimes not able to be verified by the referenced table. Additionally, these statements often required rewording to ensure that all parts of the statement can be verified by the table, which was a step taken only for the development and test sets. This left us with 981 tables and 4506 statements. The majority of the removals were due to grammatical errors as most BPO workers are not native English speakers. See Table~\ref{table:stats} (first row) for detailed statistics of the crowd-sourced training set.

\begin{figure*}[ht]
    \centering
    \includegraphics[width=\textwidth]{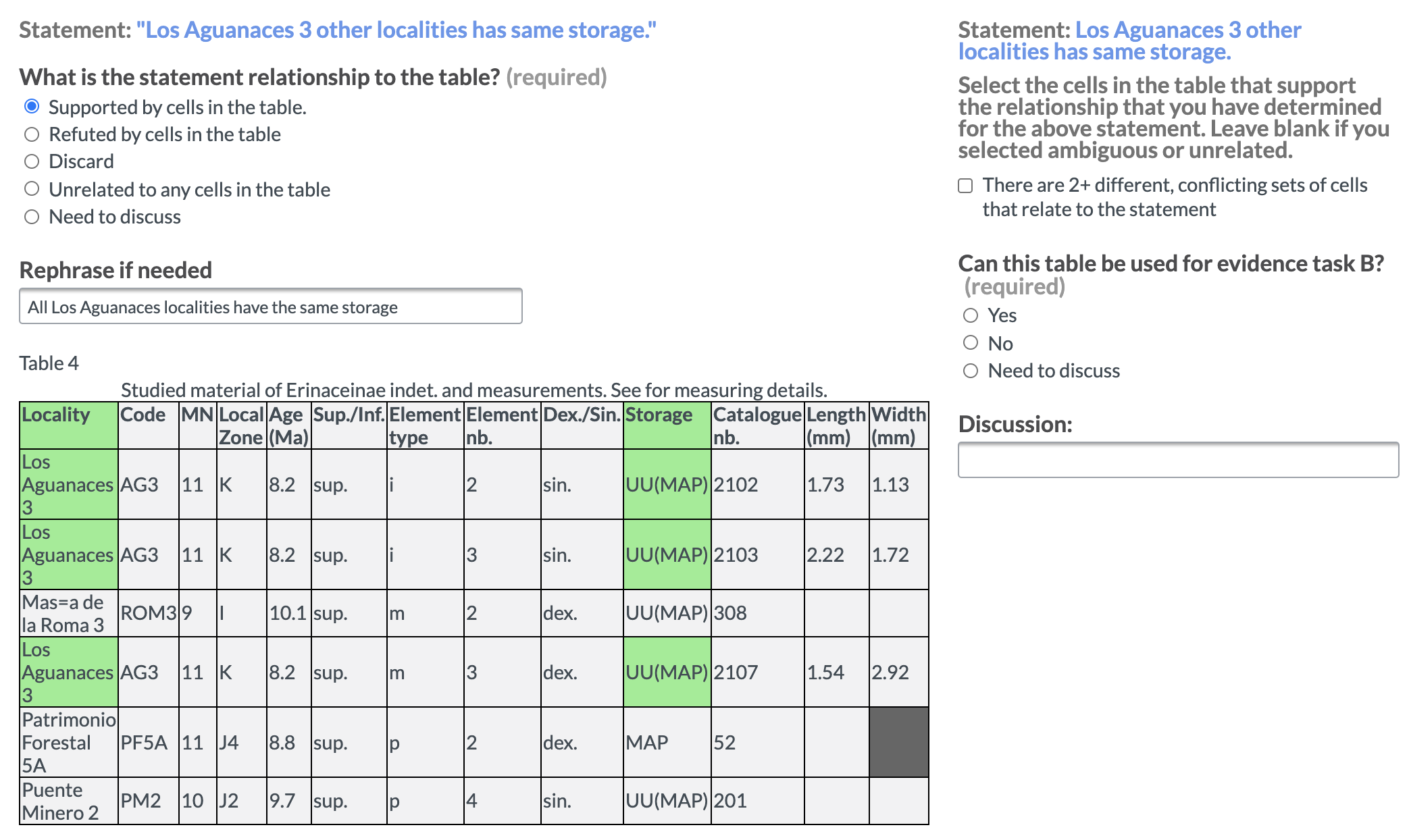}
    \caption{Screenshot showing the labeling interface for statement rephrasing, relationship labeling and evidence annotation.}
    \label{fig:screenshot}
\end{figure*}

We initially attempted to collect the development and test sets as well as evidence annotations via the same method as the training set. However, we found that the quality was not gold-level and thus we (three of the authors) decided to manually correct the statements and annotate the evidence ourselves. All authors first annotated a small set of 102 statements to test inter-annotator agreement for statement relationship and evidence labeling. Out of 102 statements, we found 5 statements where at least one of three annotators disagreed on the relationship and a further 5 statements where the relationship was agreed but the evidence annotation differed. The other 92 were in complete agreement, indicating high agreement. Therefore, the annotations for the rest of the dev set were annotated by just one person.   
The test set  was annotated fully by one author and the two other authors checked the annotations with all disagreements being resolved. See Figure \ref{fig:screenshot} for a screenshot of the statement annotation correction and evidence annotation interface. See the third and fourth rows of Table~\ref{table:stats} for detailed statistics of the dev and test sets.

\subsection{Automatically generated statements}

\begin{table*} 
\small
\centering
\begin{tabular}{p{.7cm} p{5cm} p{3cm} p{5.5cm}}
\toprule
\textbf{Input} & \textbf{Template} & \textbf{Evidence} & \textbf{Example Statements} \\
\midrule
col\_i, col\_j & `The' + col\_i\_head + `is'  + col\_i\_val + `, when the' + col\_j\_head + `is' + col\_j\_val & col\_i\_head, col\_j\_head, col\_i\_val, col\_j\_val & 
The Code is AG3 when the Locality is Los Aguanances3.
\\
\midrule
col & col\_val + `is in' + col\_head & col\_val, col\_head for entailed; col for refuted & 
AG3 is in Code.
\\
\midrule
col & unique or same values & col for entailed; None for refuted & 
Sup./Inf. has the same values.
\\
\midrule
col[\#] & `The maximum of' + col\_head +`is'+val & col[\#] for entailed; None for refuted & 
The maximum of Length(mm) is 2.22.\\
\midrule
col[\#]  & `The minimum of' + col\_head +`is'+val & col[\#] for entailed; None for refuted & 
The minimum of Length(mm) is 1.54.\\
\midrule
col[\#]  & `The mean of' + col\_head + `is' + val & col[\#] & 
The mean of Length(mm) is 1.83. \\
\midrule
col[\#]  & `The median of' + col\_head + `is' + val & col[\#] & 
The median of Length(mm) is 1.73.\\
\midrule
col[\#]  & `The mode of' + col\_head + `is' + val & col[\#]  & 
The mode of Length(mm) is 1.54, 1.73, 2.22. \\
\bottomrule
\end{tabular}
\caption{\label{table:templates} Template and evidence rules used for auto-generated ground truth. The examples are derived from Table 4 in Figure 4.}
\end{table*}

IBM Watson™ Discovery\footnote{https://www.ibm.com/cloud/watson-discovery} is an AI-powered search and text analytics engine for extracting answers from complex business documents.
One of the available functionalities is a Table Understanding service that produces a detailed enrichment of table data within an html document. We use this service to identify the body and header cells, as well as the \emph{cell relationships}, within our dataset. We then proceed to use a set of templates to automatically create statements about each table. We begin by identifying which cells and columns are numeric and non-numeric using a simple regex. Unlike non-numeric cells, numeric cells and columns are appropriate for specific templates that expect numeric values, such as `Value [V] is the maximum of Column [C]', where every value in column [C] has been identified as numeric. We also generate evidence for some of these templates. The template and evidence generation rules along with their inputs are detailed in Table \ref{table:templates}. This process generated 3,512,978 statements from 1,980 tables which were highly skewed in favor of refuted statements. This dataset was then down-sampled to a maximum of 50 statements per table while ensuring a more even distribution between the two classes to form our final released auto-generated dataset. The full statistics for the auto-generated training data is shown in the second row of Table~\ref{table:stats}.

\section{Evaluation Metrics}
\subsection{Task A: Statement Fact Verification}
The goal of task A is to determine if a statement is entailed or refuted by the given table, or whether, as is in some cases, this cannot be determined from the table.
We show two evaluation results. The first is a standard 3-way Precision / Recall / F1 micro evaluation of a multi-class classification that evaluates whether each table was classified correctly as Entailed / Refuted / Unknown. This tests whether the classification algorithm understands cases where there is insufficient information to make a determination.
The second, simpler evaluation, uses the same P/R/F1 metric but is a 2-way classification that removes statements with the ``unknown'' ground truth label from the evaluation. The 2-way metric still penalizes misclassifying refuted/ entailed statement as unknown. 

\subsection{Task B: Cell Evidence Selection}
In Task B, the goal is to determine for each cell and each statement, if the cell is within the minimum set of cells needed to provide evidence for the statement (``relevant'') or not (``irrelevant''). In other words, if the table were shown with all other cells blurred out, would this be enough for a human to reasonably determine that the table entails or refutes the statement?
The evaluation calculates the recall and precision for each cell, with ``relevant'' cells as the positive category. For some statements, there may be multiple minimal sets of cells that can be used to determine statement entailment or refusal. In such cases, our dataset contains all of these versions. We compare the prediction to each ground truth version and count the highest score.

\section{Experiments}
We present our baseline experimental setup for each task below.
\paragraph{Task A}
We employ state-of-the-art Table-BERT implementation\footnote{\url{https://github.com/wenhuchen/Table-Fact-Checking}} as proposed by \citet{chen2020tabfact}. We utilize Table-BERT's best performing configuration (Table-BERT-Horizontal-T+F-Template) as (1) using entity-linking to find the relevant columns for a statement, (2) flattening the table by scanning horizontally to form natural statements from the relevant columns and their cell values and (3) classifying the flattened table and the statement using the sentence pair classification setting in BERT. To overcome the lack of unknown statements in our dataset, we supplement each table with randomly chosen statements from other tables. In Table-BERT, if the entity linking results in no matches, the flattened table is marked as [UNK]. As our dataset contains unknown statements, in such cases we consider all columns to be a match and flatten the entire table.

Using the above process, we perform the following experiments (1) apply the Table-BERT model out-of-the-box (2) re-train Table-BERT model with unknown statement and apply on our test data (3) fine-tune the model in (2) with our manual+auto-generated data and apply on our test data. We also compare these experiments with a majority baseline with entailed as our majority class. The results are presented in Table \ref{table:resultstasksa}. Applying Table-BERT model out-of-the-box provides some improvement over a majority-baseline. However, when the model is retrained with previously missing unknown statements, the performance improves for three-way classification. Further fine-tuning the model with our training dataset (both manual and auto-generated) provides the best performance on the two-way F1-score.

\begin{table}
\centering
\small
\begin{tabular}{p{4cm}  l  l }
\toprule
\multirow{2}{*}{\textbf{Experiment}} &   \multicolumn{2}{c}{\textbf{Test}}  \\
& \textbf{2-way} & \textbf{3-way} \\
\midrule
majority-baseline  & 52.42 & 42.16\\
\midrule
original Table-BERT & 56.77 & 45.58 \\
\midrule
re-trained Table-BERT & 52.96 & 48.33 \\
\midrule
+ FT with SEM-TAB-FACTS & 56.81 & 48.24 \\
\bottomrule
\end{tabular}
\caption{\label{table:resultstasksa} Task A baseline results using F1-score.}
\end{table}

\paragraph{Task B}
We present the following two baselines for Task B: (1) a random baseline where each cell is marked relevant or irrelevant randomly (2) a simple word-match-based baseline where a cell is marked relevant if it overlaps with the statement. The baseline results are presented in Table \ref{table:resultstasksb}.

\begin{table}
\small
\centering
\begin{tabular}{lll}
\toprule
\textbf{Experiment} & \textbf{Dev} & \textbf{Test}\\
\midrule
random-baseline & 21.18 & 20.47\\
\midrule
word-match & 49.53 & 47.39\\
\bottomrule
\end{tabular}
\caption{\label{table:resultstasksb} Task B baseline results using F1-Score}
\end{table}

\begin{table}[t]
\small
\centering
\begin{tabular}{ccc}

\toprule
\textbf{Team}  &\textbf{3-way F-Score} & \textbf{2-way F-Score}  \\
\midrule
\multicolumn{3}{c}{\textbf{Official Leaderboard}}  \\
\midrule
King001&\textbf{84.48}&\textbf{88.74}\\
THiFly\_Queen&83.76&84.55 \\
RyanStark&81.51&87.22\\
sattiy&77.32&84.96\\
BreakingBERT@IITK&69.31&76.81\\
Volta&67.34&72.89\\
TAPAS&66.81&73.13\\
AttesTable&65.59&71.72\\
Yaoxu&60.76&75.8\\
Beary-group&58.37&72.56\\
ok-team&57.79&71.84\\
SUNLP&47.92&59.58\\
FishToucher&41.83&52.01\\
KaushikAcharya&36.23&23.08\\
\midrule
\multicolumn{3}{c}{\textbf{Unverified Leaderboard}}  \\
\midrule
Skywalker&92.55&95.15\\
MagicPai&90.88&94.03\\

endworld&82.35&88.16\\
Paima&81.96&88.85\\

ravikranc&57.90&71.99\\
\bottomrule
\end{tabular}
\caption{\label{table:taskaleaderboard} Task A Leaderboard}
\end{table}

\section{Competition Results}

We present two leaderboards for each task\footnote{We made the assumption that teams would not make any use of the test data, as is usually the case for algorithm evaluation, but we did not make this explicit ahead of time and some teams did not realize this was an issue. We decided to have two leaderboards to have a fair comparison for all teams.}. The official leaderboard is from participants who have given us detailed descriptions on their system and affirmed that they did not incorporate any information from the test set that changed their final model. This is a more accurate representation of system quality. The unverified leaderboard is composed of participants who either did not give enough detail or have affirmed that they incorporated some test data information in their final model. The participants did not have access to labels for test data but some teams altered their models upon examining the input data in the test set. Although we discouraged this approach, we present the results in hopes it can give some interesting information about how much improvement might be possible with having access to input test data.

19 teams participated in Task A. Of the 14 teams on the official leaderboard, King001 obtained the highest score for task A for both the 2-way (88.74) and 3-way (84.48) F-scores. However, the top three participants have comparable scores. All teams except for the last two beat our best baseline in Table~\ref{table:resultstasksa}. The unverified leaderboard includes 5 teams and contains higher scores thank in the official leaderboard. However, due to the reasons outlined above, we cannot say with certainty that the results are reproducible. The full leaderboard results for all participants are in Table \ref{table:taskaleaderboard}.

Task B is a much harder task and fewer teams participated in this challenge. Of the 12 teams that participated, 8 are in the official leaderboard. The best score is 65.17 by BreakingBERT@IITK(65.17) which is noticeably lower than the F-scores in Task A. Similarly to Task A the results in the unverified leaderboard are considerably higher. The full leaderboard results for all participants are in 
Table \ref{table:taskbleaderboard}.

\begin{table}[t]
\small
\centering
\begin{tabular}{cc}

\toprule
\textbf{Team}  & \textbf{F-Score}  \\

\midrule
\multicolumn{2}{c}{\textbf{Official Leaderboard}} \\
\midrule
BreakingBERT@IITK&\textbf{65.17}\\
Volta&62.95\\
King001&62.14\\
FishToucher&60.06\\
RyanStark&54.96\\
Sattiy&48.56\\
AttesTable&43.02\\
KaushikAcharya&33.81\\
\midrule
\multicolumn{2}{c}{\textbf{Unverified Leaderboard}}  \\
\midrule
MagicPai&88.74\\
SkyWalker&73.05\\

endworld&57.85\\
Paima&51.97\\
\bottomrule
\end{tabular}
\caption{\label{table:taskbleaderboard} Task B Leaderboard}
\end{table}


We summarize the system details for all participating teams in Tables \ref{table:techniquesA} (Task A) and \ref{table:techniquesB} (Task B). In general, deep learning was the most popular approach used  by the participants e.g. BiLSTM with attention, BERT \cite{devlin2019bert} etc. Most of the participants used transformer-based models to train their systems with flavors ranging from general-domain BERT \cite{devlin2019bert} to table-understanding specific versions like TAPAS \cite{herzig-etal-2020-tapas}, TaBERT \cite{yin-etal-2020-tabert} and Table-BERT \cite{chen2020tabfact}. One third of the participants employed some form of ensembling technique in their submission.

\begin{table*}\setlength{\tabcolsep}{3pt}
\small
\centering
\begin{tabular}{p{3cm}  p{13.3cm}}
\toprule
\textbf{Team}  &\textbf{Description}  \\
\midrule
AttesTable~\cite{varma2021attestable} & Extended TAPAS to 3 classes by fine-tuning it. Employed a novel way of synthesizing ``unknown'' samples. \\
\midrule
BreakingBERT@IITK \cite{BreakingBERT2021semeval} &
Ensemble models with TAPAS and TableBERT Transformers in a hierarchical two-step method for 3-way classification (unknown vs not unknown first) \\
\midrule
Beary-group & Used TAPAS model with TabFact task, and added unique features. Employed prepossessing tricks like k-fold validation and replacing the characters and did hyperparameter tuning. \\
\midrule
BOUN \cite{boun_semtabfact}* & Used text augmentation techniques such as back translation and synonym swapping on the TAPAS model. Domain adaptation and joint learning using SemTabFacts and TabFact datasets. \\
\midrule
endworld & Data Cleaning. Ensemble combining 80 instances of trained TaPas-Large and label smoothing. \\
\midrule
FishToucher & Motivated by TaPas, used BERT and enriched the embedding layer with two new token type embeddings: row and column ids*
\tiny{(*The team mistakenly submitted an old model version, see paper for more accurate scores)} \\
\midrule
Kaushik Acharya \cite{acharya2021semeval} & Parsed statements into candidate logical form; mapped result to handwritten rules, to then execute over relevant cells (identified using string matching and universal dependency parsing) \\
\midrule
King001 & Trained 20 instances of TaPas, SAT and Table-Bert for an ensemble of 60 models. Used preprocessing like acronym completion, rules to align the table content with the question content, label smoothing. \\
\midrule
MagicPai& Multi-model training using models such as TaBERT, tapas\_wikisql, tapas\_TabFact, tapas\_masklm. Finally rule amendments and aligning the distribution of training and test data \\
\midrule
ok-team & TAPAS pretrained on TabFact with preprocessing of data (like transforming English numerals to Arabic numerals, removing special characters etc.)  \\
\midrule
Paima & Fine-tuned TAPAS optimized to perform window scanning on statement-related table data. Pre-processing to reduce abbreviations for table headers, and identifying operation expressions. \\
\midrule
RyanStark & Multi-model TaBERT pretrained Model fusion. Pre-processing such as case and abbreviations. \\
\midrule
Sattiy~\cite{Sattiy2021semeval} & Ensemble of 6 fine-tuned pre-trained models on the augmented data with content snap-shot input. Augmented the data provided by expanding the labels. Used Fast Gradient Method and added disturbance to the embedding layer to obtain a more stable word representation and a more general model. \\
\midrule
SkyWalker & Deep learning, LPA rules, TAPAS dataset \\
\midrule
SUNLP & BERT for sequence classification, transfer learning \\
\midrule
TAPAS~\cite{Mueller2021} &  Ensemble of TAPAS (BERT-large-like) models: trained with a Mask-LM task on Wikipedia tables, intermediate pre-training data and TabFact data. Hierarchical two-step method for 3-way classification. Added neutral statements during training: random and by removing one of the evidence columns.  \\
\midrule
THiFly\_Queen~\cite{thiflyqueen2021semeval} & Ensemble models in a hierarchical two-step method. 8-model to identify unknown statements and 9-model ensemble to classify entailed/refuted. Incorporated different ensemble weights for various statement types (count, superlative, unique). \\
\midrule
Volta~\cite{SemEval2021-9-Volta} & Finetuned TAPAS that was pretrained on TabFact. Pre-processing to standardize multiple header rows to a single header.  \\
\midrule
Yaoxu & Added numeric and enumerate features to TAPAS and also statistic information (such as count) as a new row/column to the table.  \\ 
\bottomrule
\end{tabular}
\caption{Descriptions of systems from participants for Task A. *Note: Team BOUN did not participate in the official leaderboard.}
\label{table:techniquesA}
\end{table*}

\begin{table*}\setlength{\tabcolsep}{3pt}
\small
\centering
\begin{tabular}{p{2cm}  p{13.3cm}}
\toprule
\textbf{Team}  &\textbf{Description} \\
\midrule
BreakingBERT @IITK & An ensemble of an individual cell-based NLI approach and a similarity approach with the cells and statement \\
\midrule
FishToucher & BERT CLS tokens for statement and table cells are used to determine cell relationships to each other, and the statement (for relevant cells) \\
\midrule
Kaushik Acharya & Relevant cells are output as part of Task A \\
\midrule
RyanStark & BOW approach with rules applied based on word matches in header and data cells. \\
\midrule
Volta & Finetuned TAPAS for cell selection. Different models for entailed and refuted statements. Used transfer learning and header standardization. \\
\bottomrule
\end{tabular}
\caption{Descriptions of systems from participants for Task B (when provided)}
\label{table:techniquesB}
\end{table*}

Most of the participants have used the manually generated ground-truth in the development of their systems, with only one team not finding it useful. Further, a large percentage of participants have used the auto-generated ground truth in their systems with three teams not finding it helpful in their evaluation. 

In terms of external resources, a majority of the participants used external table understanding resources in their systems. Further, most of the participants employed pre-processing techniques like acronym completion, removing special characters, etc... A substantial percentage of participants used techniques like incorporating word embeddings, entity resolution etc. Finally, a large number of participants used TabFact \cite{chen2020tabfact} as an external dataset.

\begin{table}
\small
\centering
\begin{tabular}{rccc}

\toprule
& \textbf{Refuted} & \textbf{Entailed} & \textbf{Unknown} \\
\midrule
\textbf{Refuted} & 164 & 81 & 3\\
\midrule
\textbf{Entailed} & 46 & 226 & 2\\
\midrule
\textbf{Unknown} & 16 & 72 & 43\\
\bottomrule
\end{tabular}
\caption{\label{table:taskaconfusion} Task A average confusion matrix}
\end{table}

We also conducted additional analyses on participant submissions on the official leaderboard. We show through the average confusion matrix for Task A in Table \ref{table:taskaconfusion} that the Unknown label was the most difficult. In fact, there were more unknown statements incorrectly labelled as entailed than were correctly categorized. Naturally, the statements with the lowest accuracy ($< 25 \%$) consist of mainly unknown statements, especially those statements that have words overlapping with those in the table. Out of the entailed and refuted statements, ones that require numerical reasoning, like range, count or comparisons seemed to be most challenging. The statements with the highest accuracy ($> 95 \%$) generally had most words or numbers exactly overlapping with those in the table. In task B, out of the statements with less than 30\% evidence F-score, 86\% were ones with a refuted relationship. Conversely, the statements with greater than 70\% F-score, 74\% were ones with an entailed relationship. This shows that it is more difficult to find the most direct evidence to prove that a statement is refuted by a table than it is to show the positive evidence that a particular statement is supported by it. We believe this is an interesting line of research for future studies. 

\section{Conclusion and Future Works}

In this paper, we presented the data and competition results for SEM-TAB-FACTS, Shared Task 9 of SemEval 2021. We created a large dataset via automated and crowdsourced fact verification as well as evidence finding for tables. Our 19 teams had a variety of techniques to tackle this unique but very relevant problem. The evidence finding scores are still quite low and have a large improvement potential. Additionally, the test set may be expanded in future versions of this task with a combination of manually generated, natural, and automated statements. 
\bibliographystyle{acl_natbib}
\bibliography{anthology,acl2021}
\end{document}